\documentclass[10pt]{article} 

\usepackage[accepted]{rlj} 

\usepackage{amssymb}            
\usepackage{mathtools}          
\usepackage{mathrsfs}           
\usepackage{graphicx}           
\graphicspath{{figures/}}
\usepackage{subcaption}         
\usepackage[space]{grffile}     
\usepackage{url}                
\usepackage{lipsum}             
\usepackage{multirow}
\usepackage{tabularx}
\usepackage{wrapfig}
\usepackage{makecell}
\usepackage{tablefootnote}

\usepackage{Definitions}

\usepackage{booktabs}

\title{\libname: A Taxonomy and Modular Framework for Reinforcement Learning with Diffusion Policies}

\setrunningtitle{A Taxonomy and Modular Framework for Reinforcement Learning with Diffusion Policies}

\author{
Chenxiao Gao\textsuperscript{1}, Edward Chen\textsuperscript{1}, Tianyi Chen\textsuperscript{1}, Bo Dai\textsuperscript{1}
}

\emails{\{cgao,echen324,tchen667,bodai\}@gatech.edu}

\affiliations{
$^{1}$\textbf{Georgia Institute of Technology}\\
}

\contribution{
    We establish a clear categorization of modern RL algorithms with diffusion policies (\dprl), organizing them based on their regularizations and guiding mechanisms, thereby inspiring the development of more efficient and easier-to-use RL algorithms for diffusion policies.
    }
    {
    Although diverse methods have emerged recently, the lack of a unified perspective has led to fragmented research. Our taxonomy aims to identify fundamental similarities and differences across the published research. 
    }

\contribution{
    We introduce a unified open-source codebase based on JAX, leveraging JIT-compilation for high-throughput training and inference. This framework is also designed with a modular style, enabling users to swap between diffusion schedules, network architectures, and benchmarks with minimal code changes.  
    }
    {
    This addresses the reproducibility gap in the field while providing a research-friendly tool for efficient prototyping.
    }

\contribution{
    We carried out standardized benchmarking and empirical analysis using the introduced framework. We provide comprehensive benchmark results across a variety of continuous control tasks, including Gym-Locomotion, DeepMind Control Suite, and IsaacLab, offering a side-by-side comparison of diffusion-based methods, revealing the empirical properties of the existing algorithms, and thus, establishing guidelines for practitioners. 
    }
    {
    Current research in DPRL often lacks standardized evaluation protocols to truly isolate the effect of algorithm designs from confounding factors like differences in noise schedules, network architectures, hyperparameter tuning, and training setups.  
    }


\summary{
Thanks to their remarkable flexibility, diffusion models and flow models have emerged as promising candidates for policy representation. However, efficient reinforcement learning (RL) upon these policies
remains a challenge due to the lack of explicit $\log$-probabilities for vanilla policy gradient estimators.
While numerous attempts have been proposed to address this, the field lacks a unified perspective to reconcile these seemingly disparate methods, thus hampering ongoing development.
In this paper, we bridge this gap by introducing a comprehensive taxonomy for RL algorithms with diffusion/flow policies. To support reproducibility and agile prototyping, we introduce a modular, JAX-based open-source codebase that leverages JIT-compilation for high-throughput training. Finally, we provide systematic and standardized benchmarks across Gym-Locomotion, DeepMind Control Suite, and IsaacLab, offering a rigorous side-by-side comparison of diffusion-based methods and guidance for practitioners to choose proper algorithms based on the application.  Our work establishes a clear foundation for understanding and algorithm design, a high-efficiency toolkit for future research in the field, and an algorithmic guideline for practitioners in generative models and robotics. 
}

\begin{document}

\makeCover  
\maketitle  

\begin{abstract}
Thanks to their remarkable flexibility, diffusion models and flow models have emerged as promising candidates for policy representation. However, efficient reinforcement learning (RL) upon these policies remains a challenge due to the lack of explicit log-probabilities for vanilla policy gradient estimators. While numerous attempts have been proposed to address this, the field lacks a unified perspective to reconcile these seemingly disparate methods, thus hampering ongoing development. In this paper, we bridge this gap by introducing a comprehensive taxonomy for RL algorithms with diffusion/flow policies. To support reproducibility and agile prototyping, we introduce a modular, JAX-based open-source codebase that leverages JIT-compilation for high-throughput training. Finally, we provide systematic and standardized benchmarks across Gym-Locomotion, DeepMind Control Suite, and IsaacLab, offering a rigorous side-by-side comparison of diffusion-based methods and guidance for practitioners to choose proper algorithms based on the application.  Our work establishes a clear foundation for understanding and algorithm design, a high-efficiency toolkit for future research in the field, and an algorithmic guideline for practitioners in generative models and robotics. Our algorithm library is available at \href{https://github.com/typoverflow/flow-rl}{https://github.com/typoverflow/flow-rl}.

\end{abstract}


\section{Introduction}
Deep Reinforcement Learning (RL) has traditionally relied on simple distributions \citep{sac,td3,sacd}, such as the diagonal Gaussian distribution or Dirac delta distribution for policy parameterization. The appeal of such simple distributions lies in their mathematical convenience: they permit easy log-probability computation, rapid sampling, and tractable reparameterization, making them compatible with various optimization paradigms. Despite these computational benefits, simple distributions often fail to capture complex and multi-modal action distributions encountered in high-dimensional control \citep{dql,idql,sfbc}. This limitation becomes increasingly evident in recent work, when more efforts are dedicated to training generalist policies that are capable of capturing diverse human behaviors \citep{black2024pi0,kim2024openvla}. 

Recently, diffusion models (DMs) \citep{ho2020denoising,song2020score} and flow models (FMs) \citep{albergo2025stochastic,lipman2022flow} have emerged as powerful alternatives for policy representation \citep{dql,sfbc}. They both employ an iterative sampling process and therefore offer greater flexibility in distribution modeling. However, integrating DMs and FMs into the RL optimization loop is non-trivial. Traditional RL workflows rely on policy gradient \citep{ppo,trpo} or the reparameterization trick \citep{td3}, both of which are notoriously difficult for DMs and FMs \citep{song2020score}. 
While in RL, the target action distribution is typically defined implicitly through the utilities measured by value functions or return functions \citep{haarnoja2017reinforcement,peng2019advantage}, from which direct samples are not available \citep{pan2024model}, making vanilla diffusion/flow model training losses intractable.  

To address these challenges, recent literature has proposed various solutions for diffusion policy-based Reinforcement Learning (DPRL). Existing methods span disparate scenarios, including offline \citep{dql,dac,bdpo}, online \citep{qsm,dacer,sdac}, and offline-to-online RL \citep{huang2025offline}. Besides, they often involve confounding factors such as differences in noise schedules \citep{chen2023importance}, network architectures \citep{idql,dime,nauman2024bigger,lee2024simba}, and evaluation protocols. These discrepancies make it difficult to isolate the true drivers of the algorithmic performance.

Therefore, in this paper, we aim to systematize the landscape of DPRL, providing a unified perspective to study these methods. Specifically, our contributions are threefold:

1) We summarize and categorize modern DPRL algorithms based on their guidance mechanism and choice of reference policy. This taxonomy allows us to study these methods from first principles and expose the underlying mathematical relationships between them.

2) Leveraging JIT-compilation provided by JAX \citep{jax2018github} and its ecosystem \citep{deepmind2020jax,flax2020github}, we provide a modular, open-source codebase for representative DPRL algorithms with high-throughput training and inference. Furthermore, the library’s modular design allows researchers to swap environments and algorithmic components with minimal effort, significantly reducing the migration cost and barrier for RL research. 

3) Based on our systematic taxonomy and high-performance algorithm library, we conduct a large-scale comparative study of DPRL algorithms across three diverse continuous control suites: Gym-Locomotion, DeepMind Control (DMC), and IsaacLab. Our results establish rigorous baselines and provide practitioners with actionable insights tailored to specific application requirements.

\textbf{Comparison to existing surveys.}
Several surveys examine diffusion models in RL from different angles. \citet{zhu2023diffusion} and \citet{xu2025diffusion} provide broad taxonomies of diffusion roles (planners, synthesizers, policies) but lack a deep dive into the underlying policy optimization process. \citet{wolf2025diffusion} and \citet{li2025generative} shift the focus on specific applications like robotics and broader decision-making tasks. \citet{gorl} proposed a general encoder-decoder framework for online RL that can be instantiated with various algorithms; however, it lacks a broader discussion of alternative online RL approaches. The closest work to ours is \citet{choi2026review}, which reviews online diffusion policy algorithms. However, their categorization focuses only on guidance mechanisms, whereas we additionally investigate \dprl methods with different regularization objectives, thereby providing a more comprehensive overview of \dprl across online and offline settings.

\vspace{-2mm}
\section{Background}
\vspace{-2mm}
\textbf{Reinforcement Learning} is based on the framework of Markov Decision Process (MDP) $\langle\mathcal{S}, \mathcal{A}, T, R, \gamma, d_0\rangle$ \citep{sutton1998reinforcement}, where $\mathcal{S}$ is the state space, $\mathcal{A}$ is the action space, $T(s'|s, a)$ denotes the transition function, $R(s, a)$ is a bounded reward function, $\gamma$ is the discount factor and $d_0(s_0)$ denotes the initial state distribution. The agent aims to learn a policy $\pi: \mathcal{S}\rightarrow \Delta(\mathcal{A})$ to maximize the expected discounted return $\mathbb{E}_{\pi}[\sum_{t=0}^\infty\gamma^tR(s_t, a_t)]$. In online RL, the agent can interact with the environment to gather experiences for policy optimization, while in offline RL, the agent is provided with an offline dataset $\mathcal{D}=\{(s_t,a_t,s_{t+1},r_t)\}$, where $s_{t+1}\sim T(\cdot|s_t, a_t)$ and $r_t=R(s_t,a_t)\in\mathbb{R}$. We define the value functions as the expected cumulative rewards of executing a certain policy $\pi$:
\begin{equation}\label{eq:expected_q}
    \begin{aligned}
        \textstyle
        Q^\pi(s, a)=\mathbb{E}_\pi\left[\sum_{t=0}^\infty\gamma^tr_t|s_0=s, a_0=a\right], V^\pi(s)=\mathbb{E}_\pi\left[\sum_{t=0}^\infty\gamma^tr_t|s_0=s\right], 
    \end{aligned}
\end{equation}
which satisfy the Bellman equation $
    Q^\pi(s, a)=R(s, a)+\gamma \mathbb{E}_{s'\sim T}\left[V(s')\right]$.
With value functions defined, we can define the policy optimization objective \citep{wu2019behavior, geist2019theory}:
\begin{equation}\label{eq:policy_obj}
    \begin{aligned}
        \max_{\pi\in\Pi}\ \EE_\pi\left[Q^\pi(s, a)\right] - \lambda \dkl{\pi}{\nu},
    \end{aligned}
\end{equation}
where $\dkl{\pi}{\nu}$ is introduced to shape the behavior of the final policy. 
The optimal solution to \eqref{eq:policy_obj} can be obtained by the KKT conditions, \ie,
\begin{equation}\label{eq:policy_closed_form}
    \begin{aligned}
        \pi^*(a|s)\propto\nu(a|s)\exp(Q(s, a) / \lambda). 
    \end{aligned}
\end{equation}

Existing RL algorithms can be roughly categorized based on how the gradient is computed: {\bf 1),} \emph{Policy Gradient methods} \citep{sutton1999policy,trpo,ppo}, which uses the \emph{reinforce} trick to derive the gradient of \eqref{eq:policy_obj} as $\nabla_\theta \left(\EE_{\pi_\theta}\left[Q(s, a)\right] - \lambda D(\pi_\theta \|\nu)\right)=\EE_{\pi_\theta}[Q(s, a)\nabla_\theta \log \pi_\theta(a|s)]-\lambda \nabla_\theta D(\pi_\theta \|\nu)$, and {\bf 2)} \emph{Reparameterization-based methods} \citep{td3,sac}, which keeps the gradients through sampling and computes the gradients as $\EE_{\pi_\theta}[\nabla_{a_\theta}Q(s, a_\theta)\nabla_\theta a_\theta]-\lambda \nabla_\theta D(\pi_\theta \|\nu)$. 

\textbf{Diffusion Models (DMs) and Flow Models (FMs)} approximate data distributions by gradually perturbing clean data $x^0\sim q_0=p_{\rm data}$ with isotropic Gaussian noise according to a forward process \citep{song2020score, albergo2025stochastic}: 
$
x^{(t)}=\alpha_tx^{(0)} + \sigma_t\epsilon, \epsilon\sim \mathcal{N}(0, I)
$.
This emits the conditional distribution $q_{t|0}(x^{(t)}|x^{(0)})=\Ncal(x^{(t)}; \alpha_tx^{(0)}, \sigma_t^2)$. It then learns the score function $s_\theta(x^{(t)}, t)$ or velocity function $v_\theta(x^{(t)}, t)$ to reverse this process to sample from the target distribution $p_{\rm data}$. Since DMs and FMs mainly differ in the noise schedules $(\alpha_t, \sigma_t)$ and their network predictions $s_t$ and $v_t$ can be reparameterized interchangeably \citep{li2025back, lu2024simplifying}, we will discuss with DMs for simplicity, but we emphasize the proposed taxonomy is also applicable to FMs.
The target for the score function is to match the marginal score field, 
\begin{equation}
    \begin{aligned}
        s_t(\xt)=\EE_{\xzero\sim p_{0|t}(\cdot|\xt)}\left[s_{t|0}\left(\xt|\xzero\right)\right], 
    \end{aligned}
\end{equation}
where $p_{0|t}$ is the posterior distribution defined via Bayes' theorem as $p_{0|t}(x^{(0)}|x^{(t)})=q_{t|0}(x^{(t)}|x^{(0)})q_0(x^{(0)})/q_t(x^{(t)})$, and 
\begin{equation}
s_{t|0}(\xt|\xzero)=\nabla_{\xt}\log p_{t|0}(\xt|\xzero)=-(\xt-\alpha_t \xzero)/\sigma_t^2. 
\end{equation}
However, the posterior is not accessible. In practice, we employ \emph{conditional score matching}, which also yields the same target score:
\begin{equation}
    \begin{aligned}
        \Lcal(\theta)=\EE_{\xzero, \xt}\left[\|s_\theta(\xt, t)-s_{t|0}(\xt, \xzero)\|^2\right].
    \end{aligned}
\end{equation}
After training, sampling is done by solving a stochastic or ordinary differential equation (SDE/ODE). For example, 
\begin{equation}
    \begin{aligned}
        \rmd x^{(t)} = \left[ \frac{\dot{\alpha}_t}{\alpha_t} x^{(t)} - \sigma_t^2 \left( \frac{\dot{\sigma}_t}{\sigma_t} - \frac{\dot{\alpha}_t}{\alpha_t} \right) s_\theta(x^{(t)}, t) - \eta s_\theta(x^{(t)}, t) \right] \rmd t + \sqrt{2\eta} \rmd \bar{w},
    \end{aligned}
\end{equation}
where $\eta\geq 0$ controls the stochasticity and $\bar{w}$ is the Brownian motion in reverse time \citep{albergo2025stochastic, albergo2022building}. 

\textbf{Diffusion Policies} \citep{chi2025diffusion,dql,edp,dppo}  have been developed by exploiting conditional diffusion models for generating action $a$ conditioned on a given state $s$. In this paper, we use $\pi_t$ to denote the distribution of intermediate samples at diffusion time $t$ and $k$ to denote the environment time. Under this terminology, $\at_k$ denotes the intermediate action sample at $s_k$ and diffusion step $t$. 

\vspace{-2mm}
\section{Taxonomy of RL with Diffusion Policies}
\vspace{-2mm}

At its core, RL optimizes the regularized objective iteratively defined in \eqref{eq:policy_obj}, whose optimal policy takes the form $\pi(a|s)\propto \nu(a|s)\exp(Q(s, a))$, where $\nu$ is a reference policy and $Q(s, a)$ is the value function from policy evaluation. The choice of $\nu(a|s)$ determines the algorithm’s behavior across different learning paradigms:
\begin{itemize}
    \item \emph{Maximum Entropy RL}:  When $\nu=\text{Unif}(\Acal)$, the KL divergence reduces to the policy entropy up to some constant, which is a standard approach in online RL to encourage exploration; 
    \item \emph{Policy Mirror Descent}: When $\nu=\pi^{k-1}$ (the policy checkpoint from the last iteration), this ensures safe exploitation by constraining updates to the proximity of the current policy; 
    \item \emph{Behavior Regularized RL}: When $\nu=\pi_\Dcal$ (the dataset's behavior policy), this restricts the policy to the support of the offline data $\Dcal$, ensuring stable optimization in offline settings.
\end{itemize}

Consequently, categorizing RL for diffusion policies requires evaluating each method on two key questions: 
\begin{enumerate}
    \item \emph{How to effectively guide diffusion policy optimization using the value function $Q(s, a)$}, and
    \item \emph{Which reference policy $\nu$ is employed, as this determines the method's applicable scenarios.}.
\end{enumerate}
Accordingly, we examine the existing literature through the lenses of the \emph{guidance method} and \emph{reference policy}. Table~\ref{tab:summary} provides a categorization of existing DPRL methods based on our taxonomy. 

\begin{table}[htbp]

    \centering

    \renewcommand{\arraystretch}{1.3} 
    \caption{Summarization of existing DPRL algorithms, based on their guidance mechanism and reference policy.}

    \scriptsize{

    \begin{tabularx}{\textwidth}{c c X}

        \toprule

        \textbf{Guidance Method} & \textbf{Reference Policy} & \textbf{Algorithm Name} \\
        \midrule
        BoN Sampling & $\pi_{\Dcal}$ & IDQL~\citep{idql}, SfBC~\citep{sfbc} \\

        \cmidrule{1-3} 
        \multirow{2.5}{*}{Q-value Guidance} & $\text{Unif}(\Acal)$ & QSM~\citep{qsm}, iDEM~\citep{idem}, DPS~\citep{dps} \\
        \cmidrule{2-3}
        & $\pi_{\Dcal}$ & DAC~\citep{dac}, QGPO~\citep{qgpo}, Diffusion-DICE~\citep{diffusiondice} \\
        \cmidrule{1-3}
        \multirow{4}{*}{Weighted Matching} & $\text{Unif}(\Acal)$ & SDAC~\citep{sdac}, MaxEntDP \citep{dong2025maximum}, QVPO~\citep{QVPO} \\
        \cmidrule{2-3}
        & $\pi^{k-1}$ & DPMD~\citep{sdac}, FPMD~\citep{fpmd}, GeMPO~\citep{simpo} \\
        \cmidrule{2-3}
        & $\pi_{\Dcal}$ & QIPO~\citep{qipo} \\
        \cmidrule{1-3}
        \multirow{3.5}{*}{Reparameterization} & $\text{Unif}(\Acal)$ & DACER~\citep{dacer}, DACERv2~\citep{dacerv2}, DIME~\citep{dime} \\
        \cmidrule{2-3}
        & \multirow{2}{*}{$\pi_{\Dcal}$} & D-QL~\citep{dql}, BDPO~\citep{bdpo}, EDP~\citep{edp}, FQL~\citep{fql} \\
        \cmidrule{1-3}
        \multirow{2}{*}{Policy Gradient} & \multirow{2}{*}{$\pi^{k-1}$} & FPO~\citep{fpo}, FPO++~\citep{yi2026flow}, GenPO~\citep{genpo}, DPPO~\citep{dppo} \\

        \bottomrule

    \end{tabularx}

    }

    \label{tab:summary}

\end{table}

\vspace{-2mm}
\subsection{Best-of-N (BoN) Sampling}
The simplest approach is pretraining a diffusion policy to approximate the reference distribution $\nu(a|s)$ and refining it at inference time via Best-of-N sampling:
\begin{equation}
    \begin{aligned}
        a^*=\argmax_{a_i\in\{a_1, a_2, \dots, a_N\}}\ Q(s, a_i), \quad \text{where }a_i\sim \nu(\cdot|s).
    \end{aligned}
\end{equation}
This approach was first proposed in offline settings by \citet{sfbc} and \citet{idql}, where a diffusion model is trained to represent the offline data distribution and actions are subsequently refined using a learned critic $Q(s, a)$. Today, BoN sampling is increasingly used in the evaluation of modern \dprl methods. Because diffusion policies exhibit superior coverage across the action space, applying BoN sampling at inference enables the agent to select the highest-valued actions within the distribution, effectively trading additional inference-time computation for improved performance.


\vspace{-2mm}
\subsection{$Q$-value Guidance}\label{sec:qvalue_guidance}
Drawing inspiration from classifier guidance \citep{dhariwal2021diffusion}, sampling from a $Q$-weighted distribution $\pi^*(a|s)\propto \nu(a|s)\exp(Q(s, a)/\lambda)$ can be achieved by injecting \emph{action gradients} $\nabla_a Q(s, a)$ into the sampling process, due to the following score function relationship:
\begin{equation}\label{eq:qsm}
    \begin{aligned}
        \nabla_ a\log \pi^*(a|s) &= \nabla_a \log \nu(a|s) +\nabla_a Q(s, a) / \lambda.
    \end{aligned}
\end{equation}
Some algorithms utilize an explicit policy optimization step to internalize this guidance, while others inject the action gradient directly during sampling. For example, in the offline RL setting, DAC \citep{dac} performs diffusion matching over offline dataset samples combined with the action gradient; while in online RL, where $\nu = \text{Unif}(\Acal)$, QSM \citep{qsm} directly regresses the score network towards the action gradients. 

Unfortunately, directly using the action gradient is biased due to imprecise score mixing at intermediate diffusion steps. The precise score of diffusion step $t$ is given by
\begin{equation}
    \begin{aligned}
        \nabla_{a^{(t)}}\log \pi^*_t(a^{(t)}|s)=\nabla_{a^{(t)}}\log \nu_t(a|s)+\nabla_{a^{(t)}}\underbrace{\eta\log\EE_{a^{(0)}\sim p_{0|t}}[\exp(Q(s, a^{(0)})/\eta)]}_{Q_t(s, a^{(t)})}. 
    \end{aligned}
\end{equation}
QGPO \citep{qgpo} first formalizes this mismatch in offline scenarios where $\nu=\pi_\Dcal$, and introduces a contrastive learning objective to construct the intermediate $Q_t$-functions. Similarly, Diffusion-DICE \citep{diffusiondice} leverages Gumbel regression to estimate the intermediate $Q_t$-functions. In online scenarios where $\pi^*(a|s)\propto \exp(Q(s, a))$, rather than learning the $Q_t$-function explicitly, iDEM \citep{idem} and DPS \citep{dps} propose an importance-sampling approach to estimate the intermediate scores:
\begin{equation}
    \begin{aligned}
        \nabla_{a^{(t)}}\log \pi_t^*(a^{(t)}|s) &=\frac{\EE_{\hat{a}^{(0)}\sim p_{t|0}}[\exp(Q(s, \hat{a}^{(0)}))\nabla_{a^{(t)}}Q(s, \hat{a}^{(0)})]}{\EE_{\hat{a}^{(0)}\sim p_{t|0}}[\exp(Q(s, \hat{a}^{(0)}))]}
    \end{aligned}
\end{equation}
where the RHS can be approximated by Monte-Carlo samples and reweighting. The resulting smoothing of the action gradients helps alleviate the slow mixing issue in Langevin Dynamics style sampling.

\vspace{-2mm}
\subsection{Reparameterization}
\label{sec:reparameterization}

Similar to reparameterization-based methods like SAC~\citep{sac} and TD3~\citep{td3}, recent work explores reparameterizing the diffusion sampling process, and optimizing the network by maximizing $Q$-values of the generated samples. A common approach is Backpropagation Through Time (BPTT) \citep{dql,dacer,dacerv2,dime}, which computes and backpropagates the gradients across the entire sampling chain:
\begin{equation}
    \begin{aligned}
        \max_\theta  \ \EE_{a^{(0:T)}\sim \pi_\theta}[Q(s, a^{(0)}_\theta)].
    \end{aligned}
\end{equation}
However, BPTT incurs significant memory and computational overhead due to the need to preserve the computational graph across all diffusion timesteps. To mitigate this, several methods aim to amortize or bypass the sampling cost. For example, BDPO \citep{bdpo} constructs step-level value functions $Q(s, a, t)$ for each diffusion step $t$, and optimizes only for single-step transitions $\max_\theta  \ \EE_{a^{(t)}\sim \pi_{t|t+1, \theta}}[Q(s, a^{(t)}_\theta, t)]$. EDP \citep{edp} leverages the \emph{posterior mean estimate} $\hat{a}^{(0)}$ via the identity $\hat{a}^{(0)}=\EE_{a^{(0)}\sim p_{0|t}}[a^{(0)}]\approx(a^{(t)}+\sigma_t^2\epsilon_\theta)/\alpha_t$ and turns to optimize an efficient yet biased objective:\begin{equation}
    \begin{aligned}
        \max_\theta  \ \EE_{a^{(t)}, \hat{a}^{(0)}_\theta=(a^{(t)}+\sigma_t^2\epsilon_\theta)/\alpha_t}[Q(s, \hat{a}^{(0)}_\theta)].
    \end{aligned}
\end{equation}
Another line of research, exemplified by FQL~\citep{fql}, employs distillation to compress multi-step diffusion into single-step models, enabling efficient reparameterization in a single pass.

Nevertheless, a fundamental challenge for this guidance style is how to ensure the policy respects the regularization. Since the log-probability of actions is difficult to compute, these methods typically require additional loss terms for regularization. One approach is to sample from the reference distribution and apply a standard diffusion loss on these samples to anchor the policy. For example, D-QL \citep{dql} applies a score matching objective on samples from the offline dataset $\Dcal$ to prevent the diffusion policy from deviating. Alternatively, BDPO formulates the KL constraint across the entire diffusion path and enforces the constraint at every denoising step by decomposing the pathwise KL. Similarly, DIME \citep{dime} derives a tractable lower bound for action entropy, which enables optimizing the diffusion policy with the maximum entropy framework.


\vspace{-2mm}
\subsection{Weighted Matching}\label{sec:weighted_matching}

Weighted matching methods optimize diffusion policies by reframing policy improvement as a weighted supervised learning task. These methods reweight the diffusion training objective using functions derived from the $Q$-function or advantage function to bias the learned policy toward high-value actions. \citet{qipo} and \citet{sdac} demonstrate that the weighted score/flow matching objective yields an optimal policy that exactly matches the closed-form expression in \eqref{eq:policy_closed_form}, providing formal guarantees for policy improvement.

In its general form, the weighted matching objective is defined as:
\begin{equation}
\label{eq:weighted_matching_eqn}
    \min_\theta\ \mathbb{E}_{a^{(0)}, a^{(t)} \sim {\color{red} \tilde{p}_{0, t}}} \sbr{\exp\rbr{\frac{Q(s, a^{(0)})}{\lambda}} \left\| s_\theta\rbr{a^{(t)}; s, t} - s_{t|0}\rbr{a^{(t)}|a^{(0)}} \right\|^2},
\end{equation}
where $\tilde{p}_{0, t}$ represents a proposal distribution for the coupling $(a^{(0)}, a^{(t)})$, and the choice of this distribution varies by setting. In offline RL where $\nu=\pi_\Dcal$, QIPO \citep{qipo} proves that sampling $a^{(0)}\sim \Dcal, a^{(t)}\sim q_{t|0}$ recovers the optimal policy \eqref{eq:policy_closed_form}. For mirror descent where $\nu=\pi^{k-1}$, methods such as DPMD \citep{sdac} and FPMD \citep{fpmd} sample $a^{(0)}\sim \pi^{k-1}$ and $a^{(t)}\sim q_{t|0}$, which also recovers the optimal policy in \eqref{eq:policy_closed_form}. In online RL, QVPO \citep{QVPO} samples $a^{(0)}$ from a mixture of the uniform distribution and the last policy checkpoint; while SDAC \citep{sdac} and MaxEntDP \citep{dong2025maximum} concurrently propose \emph{reverse sampling}, where $a^{(0)}$ is drawn by reversing the forward perturbation kernel $q_{t|0}$ given $a^{(t)}$. Finally, GeMPO \citep{simpo} generalizes the exponential weighting in \eqref{eq:weighted_matching_eqn} to arbitrary monotone weighting functions. This not only provides a unified framework that subsumes existing algorithms as special cases, but also accommodates negative weights to encourage better exploration in practice.

\vspace{-2mm}
\subsection{Policy Gradient}

Policy gradient methods optimize the policy by estimating the gradients of the objective in \eqref{eq:policy_obj} via the policy gradient theorem. A prominent example is Proximal Policy Optimization (PPO) \citep{ppo}, which stabilizes training by applying proximal regularization relative to the previous iteration’s policy, $\pi^{k-1}$. The training objective is
\begin{equation}\label{eq:ppo}
    \max_\theta \mathbb{E}_{a\sim \pi^{k-1}}\sbr{\min\rbr{r(\theta)\hat{A}, \mathrm{clip}\rbr{r(\theta), 1-\epsilon^{\mathrm{clip}}, 1+\epsilon^{\mathrm{clip}}}\hat{A}}},
\end{equation}
where $\epsilon^{\mathrm{clip}}$ is a clipping threshold to avoid large updates and $r(\theta)=\frac{\pi_\theta\rbr{a|s}}{\pi^{k-1}\rbr{a|s}}$ is the policy ratio. 

The primary obstacle in applying policy gradient methods to diffusion policies is that computing the ratio $r(\theta)$ requires the exact log-likelihood $\log\pi(a|s)$, which is intractable due to the iterative generation process. To address this limitation, current algorithms employ various approximations of the ratio. For example, DPPO \citep{dppo} discretizes the diffusion SDE and treats the denoising process as a inner Markov Decision Process (MDP). Because each denoising step in this MDP is approximately Gaussian and the log-probability becomes tractable, this allows us to apply PPO to the individual denoising steps. GenPO \citep{genpo} constructs an invertible diffusion model by introducing dummy actions and uses the change-of-variables theorem to compute the action likelihood. However, this is computationally expensive. Finally, FPO \citep{fpo} and FPO++ \citep{yi2026flow} employs the conditional flow matching (CFM) loss as an approximation of the evidence lower bound (ELBO), and estimates the policy ratio as:
\begin{equation}\label{eq:fpo_ratio}
\hat{r}_{\mathrm{FPO}}\rbr{\theta}=\exp\rbr{\hat{L}_{\mathrm{CFM}, \theta^{k-1}}\rbr{a}-\hat{L}_{\mathrm{CFM}, \theta}\rbr{a}},
\end{equation}
where the CFM loss is estimated via Monte Carlo samples: $\hat{L}_{\mathrm{CFM},\theta}(a)=\frac{1}{N}\sum_{i=1}^{N}\|s_\theta(a^{(t)}, t)-s_{t|0}(a^{(t)}, a^{(0)})\|^2$.

\vspace{-2mm}
\section{Design of \libname}
\vspace{-2mm}
\begin{figure}[t]
  \centering
  \includegraphics[width=0.75\linewidth]{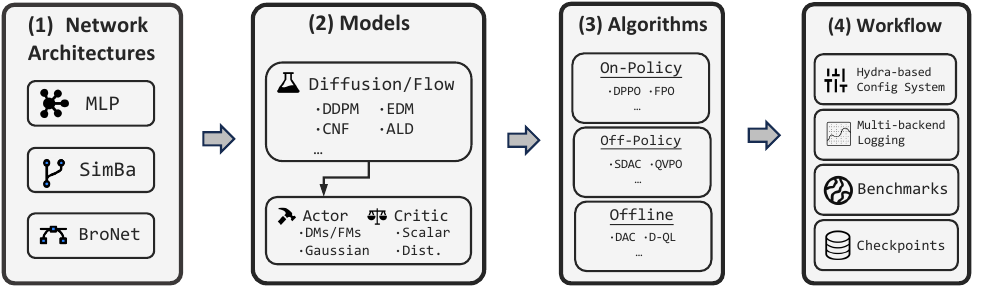}
  \caption{The overview of \libname. }
  \label{fig:teaser}
  \vspace{-2mm} 
\end{figure}

\subsection{Design Principles}
\label{sec:design_principles}

To support the large-scale comparative study and facilitate future research, our codebase, \libname, is designed according to the following principles:

\textbf{Modularity and Composability.}
The library decouples RL algorithms into orthogonal, reusable components, such as neural network architectures, actor/critic variants, generative models (flow and diffusion), and training infrastructure. Each component adheres to a shared interface, allowing researchers to combine them flexibly.

\textbf{Computational Efficiency. }\libname is built around JAX and its ecosystem. Since diffusion models require multiple iterations to sample a single action, JAX's Just-In-Time compilation provides a significant speedup in training and inference over PyTorch-based alternatives. Additionally, JAX's functional paradigm enables the native use of transformations like \texttt{vmap} (used, e.g., for critic ensembles) and \texttt{lax.scan} (for the iterative
denoising loop in diffusion).

\vspace{-2mm}
\subsection{Architecture Overview}

Figure~\ref{fig:teaser} provides a semantic overview of \libname. The system is organized into four layers, described below from bottom to top.

\paragraph{(1) Network Architecture Layer.}
While traditional RL often relies on simple backbones like Multi-Layer Perceptrons (MLPs) with ReLU activations, recent evidence suggests that RL performance scales significantly with more sophisticated architectures and increased parameter counts. Accordingly, we support a diverse suite of backbones, including standard \texttt{MLP}, as well as modern alternatives such as \texttt{SimBa}~\citep{lee2024simba} and \texttt{BroNet}~\citep{nauman2024bigger}. 

\paragraph{(2) Model Layer.} We implement DMs, FMs, and standard RL components with a common interface. For DMs, we support both discrete and continuous-time DDPM; for FMs, we support Continuous Normalizing Flows (CNFs). We also include annealed Langevin dynamics (ALD) due to its connection to DMs and FMs. The actor-critic framework supports actors parameterized by any of the aforementioned generative models or by standard deterministic and probabilistic distributions. For critics, both scalar and distributional critics are supported. All these models can utilize any architecture from the previous layer as their backbones. 

\paragraph{(3) Algorithm Layer.} Algorithms are implemented as subclasses of a \texttt{BaseAgent}. To implement a new algorithm, the user only needs to define three components: 1) the initialization logic, which assembles tools and models from previous layers; 2) the update logic, which is JIT-compiled to maximize computational throughput; and 3) sampling logic, which is greatly simplified, as the underlying DMs and FMs already provide samplers that work out-of-the-box. 

\paragraph{(4) Workflow Layer.} The workflow layer orchestrates all components through a main script. Specifically, a Hydra-based configuration system merges algorithm-specific YAMLs with shared settings, providing hierarchical hyperparameter management and command-line overrides. The main script also instantiates the environment, data buffer, and agent, then enters a loop alternating among environment interaction, dataset sampling, and compiled agent updates. The logging system supports simultaneous writes to multiple backends, including TensorBoard, Weights \& Biases, and CSV files, while Orbax handles checkpointing. Finally, we support a diverse range of tasks, spanning 2D visualization datasets, classic locomotion benchmarks such as Gym-MuJoCo and the DeepMind Control Suite, high-dimensional control tasks like HumanoidBench, and IsaacLab, which enables large-scale, hardware-accelerated simulation across multiple embodiments.

\vspace{-2mm}
\section{Benchmarking RL with Diffusion Policies}
\vspace{-2mm}
\begin{figure}[t]
  \centering
  \includegraphics[width=0.9\linewidth]{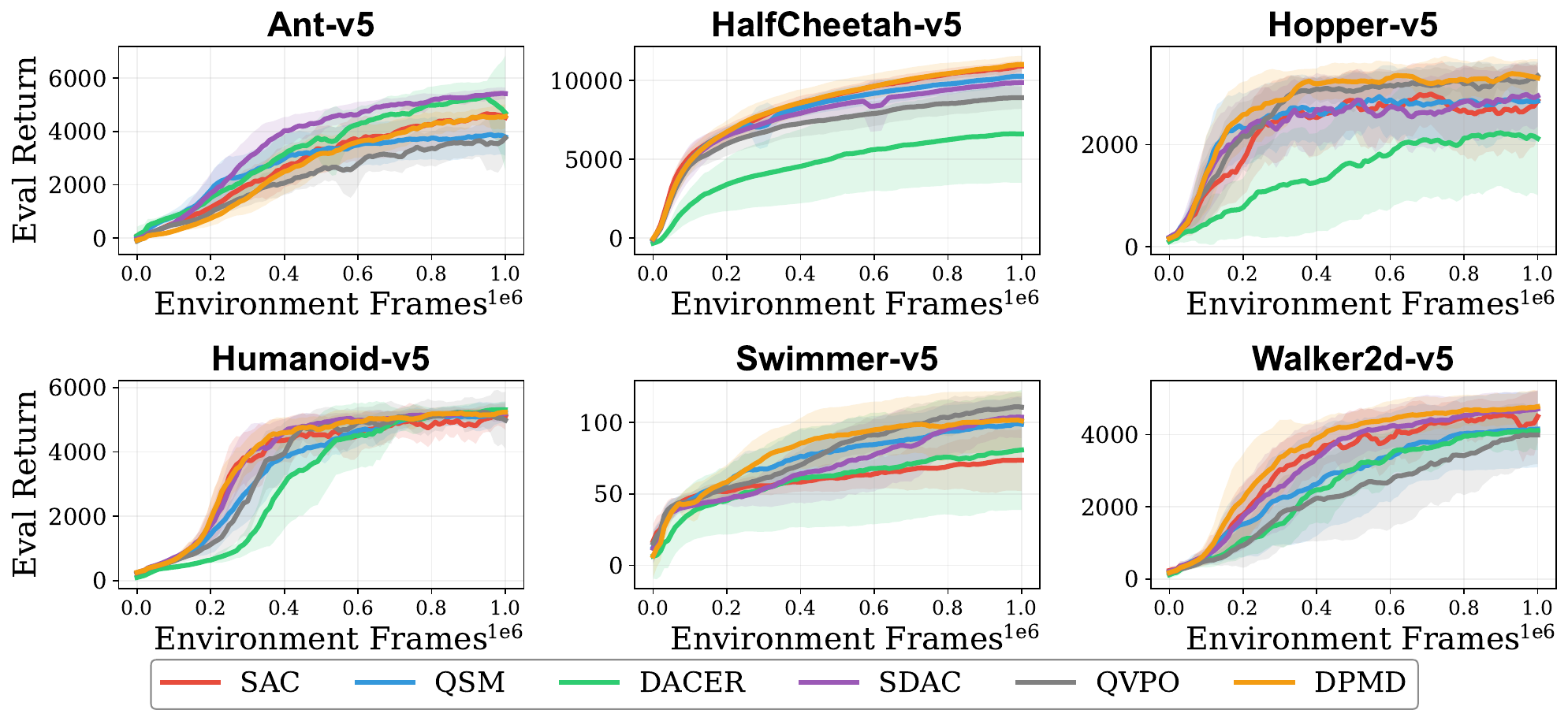}
  \caption{Training curves of several off-policy \dprl algorithms across Gym-Locomotion tasks. }
  \label{fig:general_mujoco}
  \vspace{-3mm} 
\end{figure}

\begin{figure}[t]
  \centering
  \includegraphics[width=\linewidth]{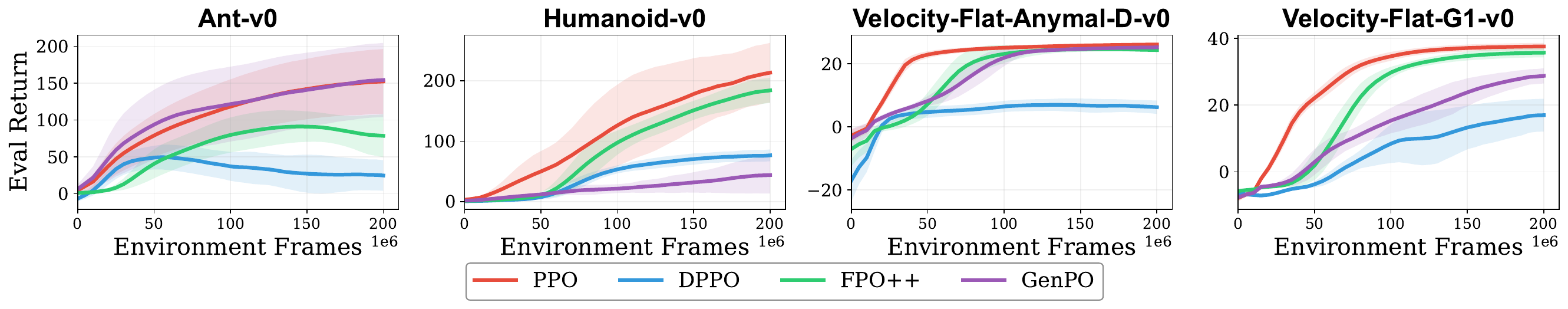}
  \caption{Training curves of on-policy \dprl algorithms on IsaacLab tasks.}
  \label{fig:general_isaaclab}
  \vspace{-2mm} 
\end{figure}

\subsection{Experiment Setup}


\newcommand{\imgw}{0.14\linewidth} 
\newcommand{\imgsep}{\hspace{2pt}}  


\paragraph{Benchmarks. }We select three complementary continuous control benchmarks that collectively span a wide range of task complexities and action dimensionalities. Gym-Locomotion \citep{gym} and DeepMind Control Suite (DMC) \citep{dmc} are both powered by the MuJoCo physics engine \citep{Todorov2012MuJoCoAP} and provide standard locomotion and balancing tasks. IsaacLab \citep{isaaclab} provides GPU-accelerated robotic environments that support massively parallel simulation across diverse embodiments, including locomotion and manipulation tasks. We use Gym-Locomotion and DMC for off-policy and offline settings due to their popularity in the literature, and IsaacLab for on-policy settings since it supports simulation at scale. The full list of tasks is provided in the Appendix~\ref{app:tasks}.

\paragraph{Algorithms. }
We evaluate representative \dprl methods from each guidance category in our taxonomy against strong Gaussian baselines. For online off-policy RL, we compare QSM \citep{qsm}, DACER  \citep{dacer}, DPMD \citep{sdac}, SDAC \citep{sdac}, and QVPO \citep{QVPO} against SAC \citep{sac}. For online on-policy RL, we compare DPPO \citep{dppo}, GenPO \citep{genpo}, and FPO++ \citep{yi2026flow} against PPO \citep{ppo}. For offline RL, we compare Diffusion-QL \citep{dql}, FQL \citep{fql}, DAC \citep{dac}, and BDPO \citep{bdpo} against IQL \citep{idql}. We also include reference scores of IDQL \citep{idql}, QGPO \citep{qgpo}, EDP \citep{edp} as a comparison. 

\begin{wrapfigure}{r}{0.35\linewidth}
  \vspace{-4mm}
  \centering
  \includegraphics[width=\linewidth]{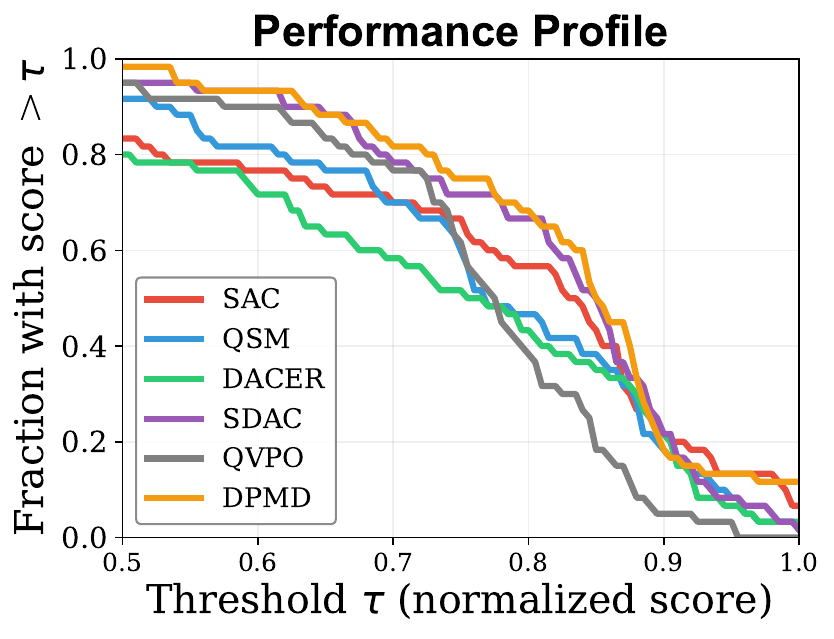}
  \caption{Performance profile on Gym-Locomotion tasks.}
  \label{fig:mujoco_tau}
  \vspace{-4mm}
\end{wrapfigure}

\paragraph{Evaluation Setup. }
To ensure a fair comparison, all methods within each benchmark follow a standardized workflow. For offline settings, we keep the hyperparameters and network designs identical to their original papers. We use the D4RL datasets for training and report the average performance of the last policy checkpoint. For the online off-policy settings, each method is trained for 1M environment frames and evaluated every 10K frames across a span of 10 episodes. No observation or reward normalization is applied in this setting. Unless an algorithm explicitly designates a specific architectural choice as a core contribution, we align all hyperparameters, network structures, and diffusion model implementations across baselines. For IsaacLab, the training pipeline follows the standard PPO rollout pipeline, utilizing 4096 parallel simulations for a total of 200M frames. For all environments, we report the undiscounted episodic return, visualized using the mean (solid line) and one standard deviation (shaded region). Detailed per-algorithm hyperparameters are provided in the Appendix~\ref{app:hyperparams}. Unless otherwise noted, all experiments are averaged over ten independent random seeds, and each evaluation uses 10 episodes.



\vspace{-2mm}
\subsection{General Performance}

\paragraph{Gym-Locomotion. }

Learning curves for the Gym-Locomotion environments are shown in Figure~\ref{fig:general_mujoco}. In Figure~\ref{fig:mujoco_tau}, we further present the performance profiles of each algorithm by normalizing the evaluation returns by per-environment reference maxima (see Table~\ref{tab:max_returns}) in Appendix~\ref{app:supplementary} and computing the fraction of runs that achieve performance above different thresholds $\tau$. Comparing the area under the curve in Figure~\ref{fig:mujoco_tau}, SDAC and DPMD deliver the strongest overall results and outperform the Gaussian baseline SAC. That said, no single \dprl algorithm dominates across all environments, and SAC remains competitive on the majority of tasks. QSM and QVPO underperform and show noticeable instability during training.

\begin{table*}[h!]
\centering
\setlength{\tabcolsep}{1.1mm}{}
\caption{Normalized returns of the final policy checkpoint for offline \dprl methods. Methods marked with $^\dagger$ denote results reported in the original paper. Numbers within 95\% of the highest scores are bolded. }
\label{tab:d4rl}
\small{
\begin{tabular}{lccccccc}
\toprule
\textbf{Dataset} & \textbf{IQL} & \textbf{IDQL}$^\dagger$ & \textbf{QGPO}$^\dagger$ & \textbf{QIPO}$^\dagger$ &\textbf{Diffusion-QL} & \textbf{DAC} & \textbf{BDPO}\\
\midrule
halfcheetah-m & 47.4 & 51.0 & 54.1 & 54.2$_{\pm1.3}$ & 50.7$_{\pm0.7}$ & 59.1$_{\pm0.6}$ & \textbf{71.2}$_{\pm0.9}$\\
hopper-m & 66.3 & 65.4 & 98.0 & 94.0$_{\pm13.3}$ & 80.4$_{\pm15.7}$ & \textbf{103.5}$_{\pm0.3}$ & \textbf{100.6}$_{\pm0.7}$\\
walker2d-m & 78.3 & 82.5 & 86.0 & 87.6$_{\pm1.5}$ & 86.7$_{\pm1.6}$ & \textbf{97.9}$_{\pm0.9}$ & \textbf{93.4}$_{\pm0.5}$\\
halfcheetah-m-r & 44.2 & 45.9 & 47.6 & 48.0$_{\pm0.8}$ & 47.4$_{\pm0.6}$ & 55.4$_{\pm0.5}$ & \textbf{58.9}$_{\pm0.9}$\\
hopper-m-r & 94.7 & 92.1 & 96.9 & \textbf{101.3}$_{\pm2.2}$ & \textbf{101.0}$_{\pm0.3}$ & \textbf{103.1}$_{\pm0.1}$ & \textbf{101.4}$_{\pm0.5}$\\
walker2d-m-r & 73.9 & 85.1 & 84.4 & 75.6$_{\pm25.1}$ & \textbf{95.7}$_{\pm1.4}$ & \textbf{98.4}$_{\pm0.5}$ & \textbf{95.5}$_{\pm1.6}$\\
halfcheetah-m-e & 86.7 & 95.9 & 93.5 & 94.5$_{\pm0.5}$ & 96.0$_{\pm0.7}$ & 100.1$_{\pm0.7}$ & \textbf{108.7}$_{\pm0.9}$\\
hopper-m-e & 91.5 & \textbf{108.6} & \textbf{108.0} & \textbf{108.0}$_{\pm5.2}$ & \textbf{106.9}$_{\pm11.7}$ & \textbf{112.3}$_{\pm1.0}$ & \textbf{111.3}$_{\pm0.2}$\\
walker2d-m-e & 109.6 & \textbf{112.7} & \textbf{110.7} & \textbf{110.9}$_{\pm1.0}$ & 108.7$_{\pm0.2}$ & \textbf{115.3}$_{\pm7.5}$ & \textbf{115.6}$_{\pm0.4}$\\
\bottomrule
\end{tabular}
}
\end{table*}

\paragraph{IsaacLab.}
In Figure \ref{fig:general_isaaclab}, we plot the return curves of three on-policy \dprl methods together with the baseline method, PPO. Overall, PPO delivers the best and most stable performance across the four tasks. Among the \dprl methods, GenPO and FPO++ deliver reasonable performances. However, we found that GenPO's training cost is significantly higher than that of PPO due to its Jacobian computation. We also observe that FPO++ occasionally collapses during training. After careful inspection, we find that this instability arises because, for samples with negative advantage, optimizing \eqref{eq:ppo} with \eqref{eq:fpo_ratio} effectively leads to an unbounded optimization problem.

\begin{figure}[t]
  \centering
  \includegraphics[width=0.85\linewidth]{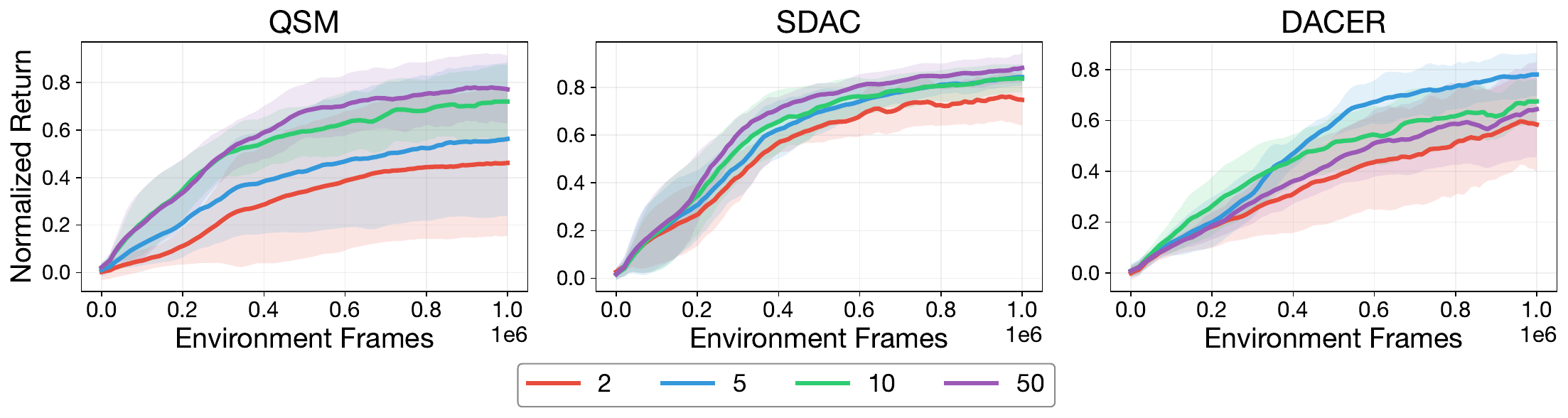}
  \caption{\textbf{Effect of diffusion steps on performance.} Performance of QSM, SDAC, and DACER when varying the number of diffusion steps (2, 5, 10, and 50).
  Returns are normalized by task-specific reference scores in Table~\ref{tab:max_returns} and averaged across Ant, HalfCheetah, Humanoid, and Walker2d.}
  \label{fig:diffusion_steps}
  \vspace{-2mm} 
\end{figure}

\begin{figure}[t]
  \centering
  \includegraphics[width=\linewidth]{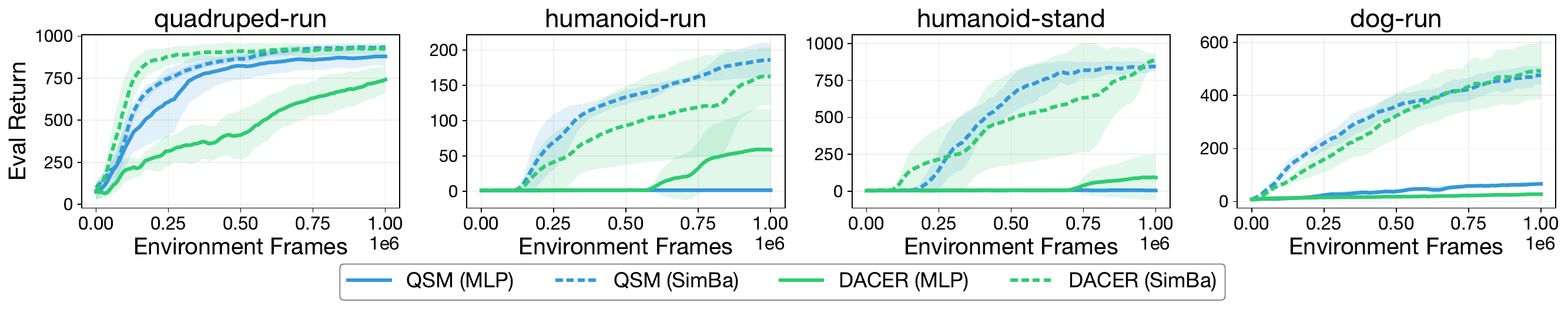}
  \caption{\textbf{Effect of network architecture.} Comparison between MLP and SimBa backbones for QSM (blue) and DACER (green) on DMControl hard tasks. Solid lines denote MLP policies and dashed lines denote Simba policies.}
  \label{fig:mlp_vs_simba}
  \vspace{-2mm} 
\end{figure}

\paragraph{D4RL. }In Table \ref{tab:d4rl}, we find that \dprl methods, especially those leveraging RL to train the diffusion policy, significantly outperform traditional offline RL algorithms such as IQL and inference-time refinement methods such as IDQL. These results demonstrate the efficacy of diffusion models not only as expressive priors for capturing complex, multi-modal offline distributions, but also as promising policy representations for offline RL.

\vspace{-2mm}
\subsection{Empirical Analysis}

\vspace{-2mm}
\paragraph{Effect of Action Dimensionality.}

\begin{wrapfigure}{r}{0.5\linewidth}
  \vspace{-4mm}
  \centering
  \includegraphics[width=\linewidth]{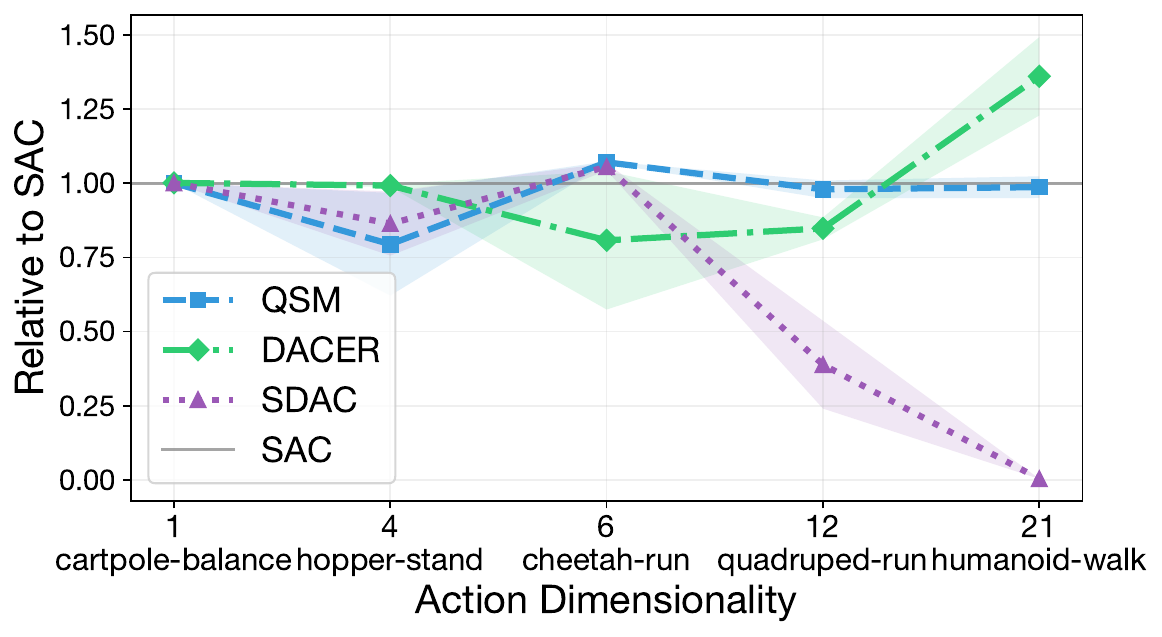}
  \caption{\textbf{Effect of Action Dimensionality.} Performance is normalized by SAC across tasks of increasing action dimension.}
  \label{fig:task_dimension}
  \vspace{-4mm}
\end{wrapfigure}

A key distinction among different guidance methods is how each utilizes the $Q$-function: both $Q$-value guidance and reparameterization methods exploit first-order gradients $\nabla_a Q$ for optimization, while weighted matching methods rely on function evaluations $Q(s, a)$ to reweight sampled actions, which is less efficient and prone to higher variance in high-dimensional action spaces. In Figure~\ref{fig:task_dimension}, we select three representative methods, QSM (from Sec \ref{sec:qvalue_guidance}), DACER (from Sec \ref{sec:reparameterization}), and SDAC (from Sec \ref{sec:weighted_matching}), and compare their performance relative to SAC with tasks of increasing action dimensionality. We find that while QSM and DACER remain competitive, the performance of SDAC degrades significantly as dimensionality grows. This suggests that for weighted matching methods to remain effective, the proposal distribution must be meticulously designed to ensure high-valued actions can be sampled and reinforced during training. 

\vspace{-2mm}
\paragraph{Effect of Diffusion Steps.}
Since diffusion models refine samples through an iterative denoising process, a natural question is how the number of diffusion steps affects each algorithm. We again select QSM, DACER and SDAC and vary the number of diffusion steps in Figure~\ref{fig:diffusion_steps}. We discover that both QSM and SDAC improve monotonically with more diffusion steps, while SDAC works significantly better even with fewer diffusion steps. In contrast, DACER exhibits the opposite trend, with its performance degrading when the number of steps exceeds five. We attribute this behavior to the nature of reparameterization-based methods, which rely on techniques such as BPTT to optimize the diffusion policy. Increasing the number of diffusion steps complicates the optimization problem, making the training process more indirect and susceptible.

\begin{figure}[t]
  \centering
  \includegraphics[width=0.85\linewidth]{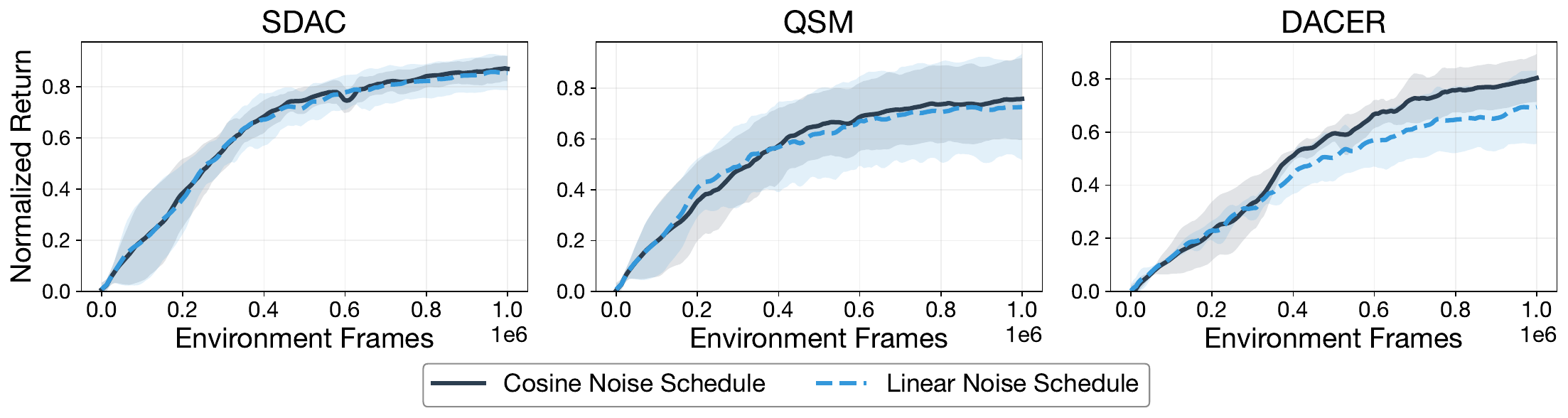}
  \caption{\textbf{Effect of noise schedule.} Performance of SDAC, QSM, and DACER with a cosine (black) or linear noise schedule. }
  \label{fig:solver-noise}
  \vspace{-2mm} 
\end{figure}


\vspace{-2mm}
\paragraph{Effect of Network Backbone.}

Recent studies \citep{lee2024simba,nauman2024bigger} have demonstrated that network architecture is a significant confounding factor in online RL performance, yet their effects are often left unexamined in the existing literature. The modular design of \libname enables users to seamlessly swap between neural network backbones. In Figure \ref{fig:mlp_vs_simba}, we select the most difficult tasks from DMC and change the network architecture from MLP to SimBa. We find that all evaluated algorithms achieve significant performance gains on tasks previously considered challenging. These results highlight the importance of controlling for architecture when evaluating algorithmic improvements.

\vspace{-2mm}
\paragraph{Effect of Noise Schedule.}

Prior work on DMs has shown that the choice of noise schedules can affect sample quality \citep{ho2020denoising,song2022denoisingdiffusionimplicitmodels}. We investigate whether these design choices carry over to the RL setting by replacing the default cosine schedule with a linear schedule. As shown in Figure~\ref{fig:solver-noise}, switching to a linear noise schedule has minimal impact on performance across all three methods. 

\begin{table}[t]
\centering
\caption{\textbf{Efficiency Comparison.} Training and inference speed of off-policy algorithms.
Training throughput is measured in gradient steps per second;
inference throughput in single-action inferences per second.}
\label{tab:speed}
\begin{tabular}{ccc}
\toprule
Algorithm & Training (gradient steps/s) & Inference (actions/s) \\
\midrule
SAC   & 1242.3 & 2623.0 \\
QSM   &  675.9 & 1285.6 \\
DACER &  603.3 &  728.3 \\
SDAC  &  256.3 & 1219.6 \\
DPMD  &  126.2 & 1313.3 \\
QVPO  &  150.6 & 1397.4 \\
QVPO (PyTorch) & 75.3 & 92.2\\
\bottomrule
\end{tabular}
\end{table}

\vspace{-2mm}
\paragraph{Efficiency Comparison.}
Built on the JAX ecosystem, \libname leverages JIT compilation to maximize computational acceleration for both training and model inference. To demonstrate this benefit and benchmark the efficiency of each algorithm, Table~\ref{tab:speed} compares their training and inference speeds. We evaluate the algorithms using \texttt{HalfCheetah-v5}, measuring training throughput in gradient steps per second and inference throughput in single-action inferences per second. All benchmarking is conducted on a dedicated NVIDIA L40S GPU. We also include benchmarks from the official PyTorch implementation of QVPO. Comparing the PyTorch and \libname versions of QVPO, we observe that our JAX-based implementation achieves approximately a 2x and 15x speedup in training and inference, respectively, making \libname a highly efficient tool for rapid algorithm development and testing. Notably, all diffusion- and flow-based algorithms impose significant computational overhead compared to non-diffusion baselines (SAC and PPO). This is primarily due to the iterative sampling required by flow policies; even during value function updates, these policies rely on an iterative sampling chain to perform temporal difference learning. Additionally, weighted matching algorithms exhibit the longest training times, primarily because they require sampling a substantial number of actions during both policy evaluation and improvement.

\vspace{-2mm}
\section{Closing Remarks}
\vspace{-2mm}
In this paper, we presented a systematic study of diffusion policy-based Reinforcement Learning. We first proposed a unified taxonomy that categorizes existing DPRL algorithms along two axes: diffusion policy optimization guidance and regularization objective. Building upon this taxonomy, we developed a modular, JAX-based library that enables high-throughput training and fair comparison of DPRL algorithms. Our standardized benchmark results across Gym-Locomotion, DeepMind Control Suite, and IsaacLab provide rigorous performance references and practical guidelines for selecting algorithm configurations. Important open questions remain in the field of DPRL, including designing algorithms robust to diverse environment characteristics, scaling to long-horizon and sparse-reward tasks, and developing a thorough understanding of the diffusion policy optimization landscape. We believe our proposed taxonomy and library provide a solid foundation for future research and practical applications in this direction.

\subsubsection*{Acknowledgments}
\label{sec:ack}
This work was supported in part by the ONR grant N000142512173, NSF grants ECCS: 2401391 and
IIS: 2403240, Dolby support, and computing resources received from the National Supercomputing
Center (CSCS) and the Swiss AI initiative.



\bibliography{main}
\bibliographystyle{rlj}

\beginSupplementaryMaterials

\section{Experimental Details}
\label{app:hyperparams}

\begin{figure}[h]
\centering
\setlength{\tabcolsep}{0pt}

\begin{tabular}{@{}cc@{\hspace{10pt}}cc@{\hspace{10pt}}cc@{}}
\multicolumn{2}{c}{\footnotesize\textbf{MuJoCo (Gymnasium)}} &
\multicolumn{2}{c}{\footnotesize\textbf{DMC}} &
\multicolumn{2}{c}{\footnotesize\textbf{IsaacLab}} \\[2pt]

\includegraphics[width=\imgw]{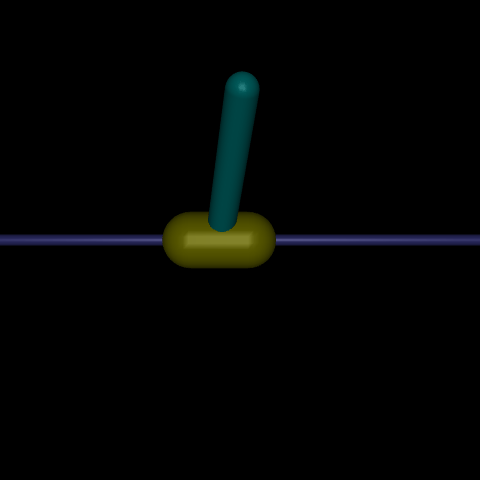} &
\includegraphics[width=\imgw]{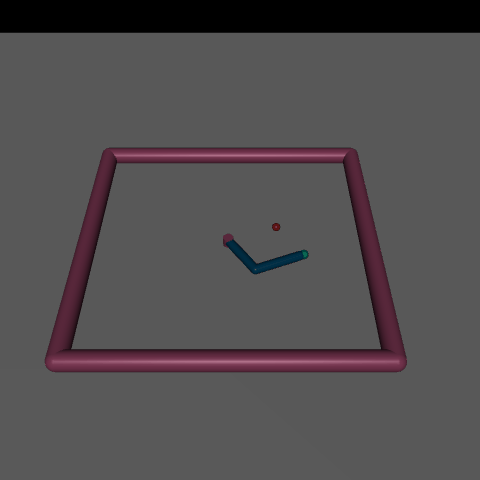} &
\includegraphics[width=\imgw]{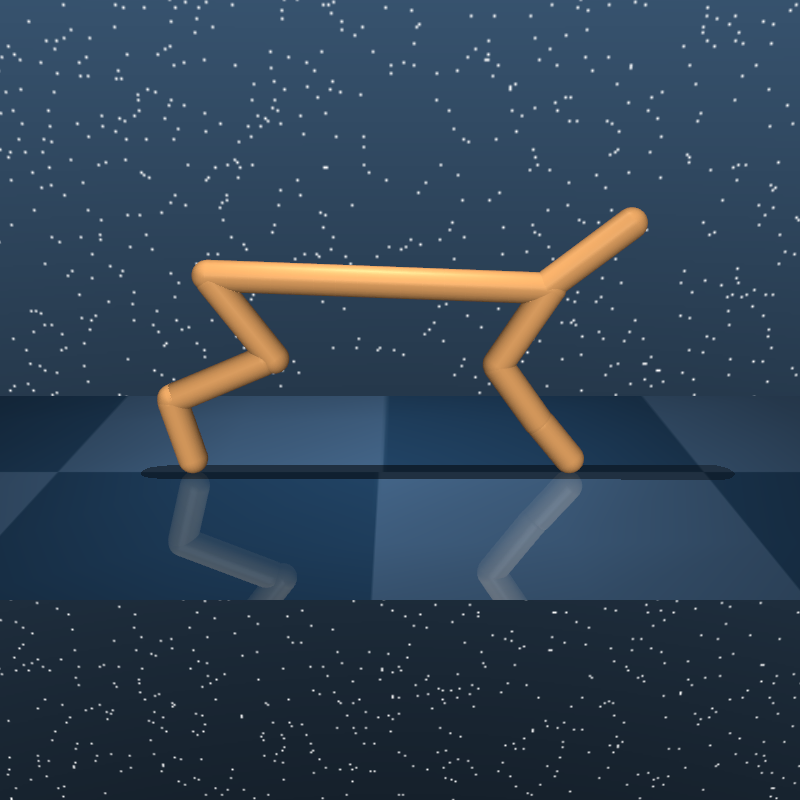} &
\includegraphics[width=\imgw]{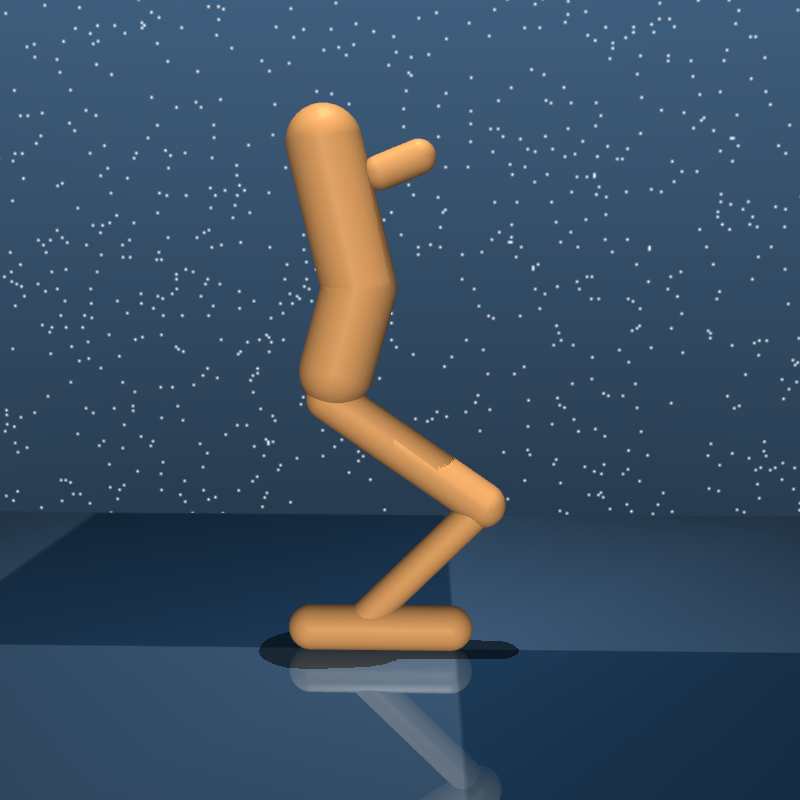} &
\isaacsq{benchmarks/isaac-ant-v0.jpg} &
\isaacsq{benchmarks/isaac-humanoid-v0.jpg} \\[-2pt]

{\scriptsize InvertedPendulum} &
{\scriptsize Reacher} &
{\scriptsize Cheetah-run} &
{\scriptsize Hopper-Stand} &
{\scriptsize Ant} &
{\scriptsize Humanoid} \\[3pt]

\includegraphics[width=\imgw]{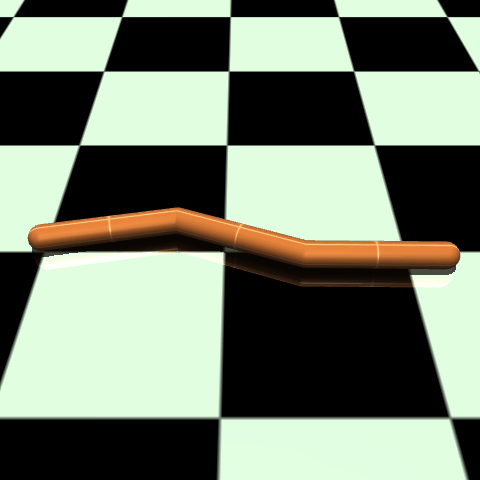} &
\includegraphics[width=\imgw]{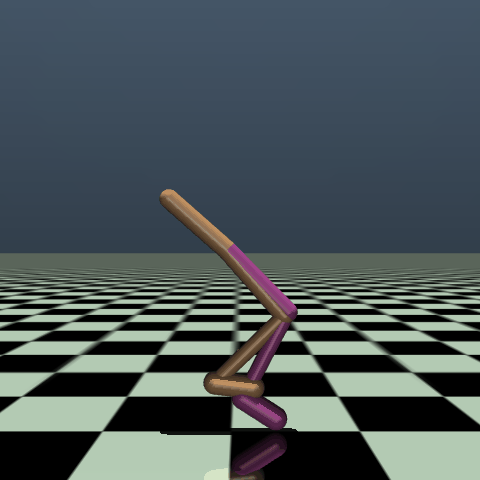} &
\includegraphics[width=\imgw]{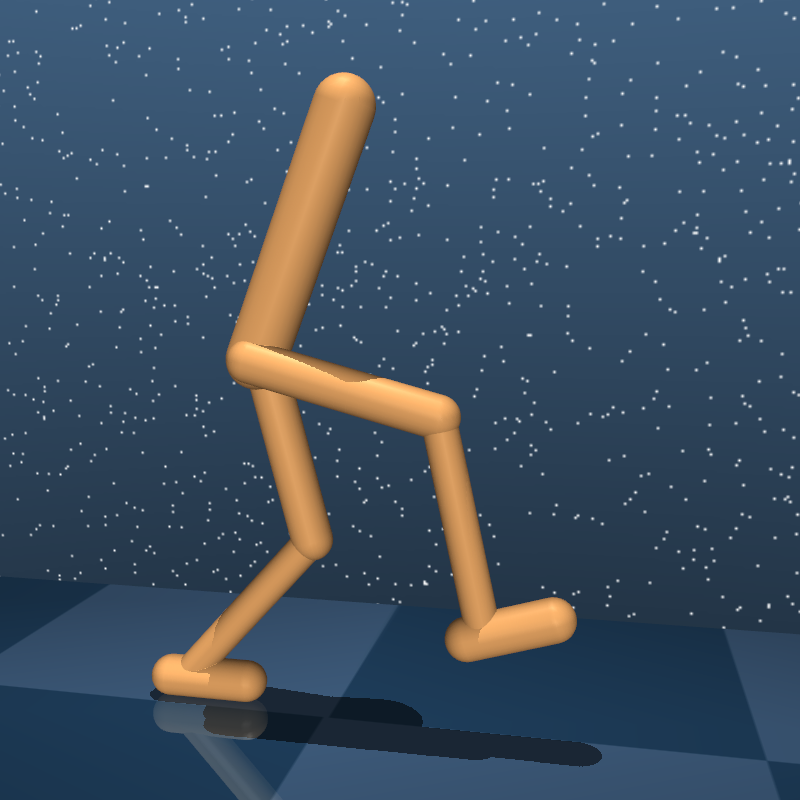} &
\includegraphics[width=\imgw]{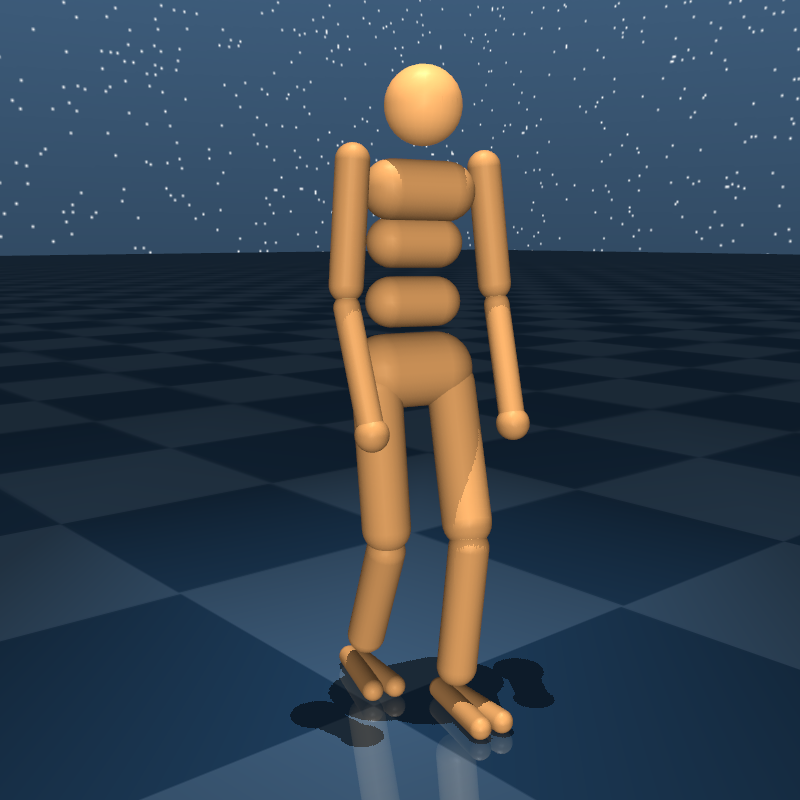} &
\isaacsq{benchmarks/isaac-lift-cube-franka.jpg} &
\isaacsq{benchmarks/isaac-velocity-flat-anymal-d.jpg} \\[-2pt]

{\scriptsize Swimmer} &
{\scriptsize Walker2d} &
{\scriptsize Walker-walk} &
{\scriptsize Humanoid-Walk} &
{\scriptsize Lift-Cube-Franka} &
{\scriptsize Velocity-Flat-Anymal-D} \\
\end{tabular}

\caption{Representative tasks across three continuous-control suites.
Left: MuJoCo Gymnasium.
Middle: DeepMind Control Suite.
Right: IsaacLab. }
\label{fig:benchmarks_overview_grid}
\end{figure}

\subsection{Benchmark Tasks}
\label{app:tasks}

We evaluate on the following continuous control tasks across three benchmarks. The illustration of representative tasks can be found in Figure~\ref{fig:benchmarks_overview_grid}. 

\paragraph{Gym-Locomotion (6 tasks).}
Ant-v5, HalfCheetah-v5, Hopper-v5, Humanoid-v5, Swimmer-v5, Walker2d-v5. 

\paragraph{DeepMind Control Suite (8 tasks).}
Cartpole-Balance, Hopper-Stand, Cheetah-Run, Quadruped-Run, Humanoid-Walk, Humanoid-Run, Humanoid-Stand, Dog-Run.  

\paragraph{IsaacLab (4 tasks).}
Isaac-Ant-v0, Isaac-Humanoid-v0, Isaac-Lift-Cube-Franka-v0, Isaac-Velocity-Flat-Anymal-D-v0. 

\subsection{Shared Training Settings}
\label{appsec:shared_setup}

\begin{table}[h!]
\centering
\caption{\textbf{Shared training settings per benchmark.}}
\label{tab:shared_hyperparams}
\begin{tabular}{lccc}
\toprule
\textbf{Setting} & \textbf{Gym-Locomotion} & \textbf{DMC} & \textbf{IsaacLab} \\
\midrule
Training frames        & 1M    & 1M    & 200M  \\
Batch size             & 256   & 512   & 24576 \\
Replay buffer size     & $10^6$ & $10^6$ & --- \\
Discount $\gamma$      & 0.99  & 0.99  & 0.99 \\
Warmup frames          & 5K    & 10K   & --- \\
Random frames          & 5K    & 10K   & --- \\
Eval frequency (frames)& 10K   & 10K   & 5M \\
Eval episodes          & 10    & 10    & 10 \\
Observation norm.      & No    & No    & Yes \\
Reward norm.           & No    & No    & No \\
Parallel envs          & 1     & 1     & 4096 \\
Frame skip             & 1     & 2     & --- \\
\bottomrule
\end{tabular}
\end{table}


Table~\ref{tab:shared_hyperparams} documents the shared training settings for each of the benchmarks. 

\paragraph{Shared off-policy defaults.}
Unless stated otherwise, all off-policy methods share a common setup. The critic is a
2-layer MLP (256 units, ReLU) on Gym-Locomotion and a deeper 3-layer MLP (512 units, ELU)
on DMC, in both cases an ensemble of 2 critics with target networks updated by an EMA of
$0.005$. The diffusion actor is a 3-layer MLP—256 units on Gym, 512 on DMC—with Mish
activations and a 64-dimensional learnable Fourier time embedding; we sample over 20 DDPM
steps under a cosine noise schedule and clip actions to $[-1, 1]$. Our SAC baseline uses a
standard Gaussian actor with tanh squashing and automatic entropy tuning. For training, we train each algorithm with 1M environment frames with frame skip of 1 (for Gym tasks) or 2 (for DMC tasks). The replay buffer size is fixed as 1M and algorithms are trained with data batches of 256 samples. 

\paragraph{Shared on-policy defaults.}
Unless otherwise specified, all on-policy methods share a standardized configuration. Both
the \textit{actor} and \textit{critic} are 3-layer MLPs with hidden dimensions
$(256, 256, 256)$ and ELU activations; the critic uses a learning rate of $10^{-3}$ and the
actor $10^{-4}$. For PPO-specific terms we use Generalized Advantage Estimation (GAE) with
$\lambda = 0.95$, a clipping parameter $\epsilon = 0.2$, and a gradient-clipping norm of
$1.0$. Each update performs $5$ epochs over $4$ minibatches with advantage normalization.
The infrastructure runs $4096$ parallel environments with a rollout length of $24$, yielding
a total batch size of $98{,}304$ per update. Note that due to the significant GPU memory cost of GenPO, we used a batch size of 6144 for GenPO. 

\subsection{Algorithm-Specific Hyperparameters}
\label{app:supplementary}


  \paragraph{Off-policy algorithm-specific hyperparameters.}
  All off-policy diffusion methods use actor and critic learning rates of $3\times10^{-4}$,
  except DACER ($10^{-4}$ for both) and DPMD ($10^{-4}$ actor, $3\times10^{-4}$ critic). QVPO
  and SAC use a 2-layer $(256, 256)$ actor; all other diffusion methods use a 3-layer
  $(256, 256, 256)$ actor.

  \textit{QSM} samples \texttt{num\_samples}=10 actions per state for the score-matching update,
  with temperature $\lambda=0.1$.

  \textit{DACER} updates the actor every 2 critic steps. It uses \texttt{entropy\_num\_samples}=200
  for Monte Carlo entropy estimation, \texttt{noise\_scaler}=0.1 (0.15 for HalfCheetah/Humanoid)
  for exploration, \texttt{reward\_scale}=0.2 (1.0 on DMC), and updates $\alpha$ every 10K steps
  with \texttt{alpha\_lr}=0.03.

  \textit{DPMD} performs a KL-constrained mirror-descent update: for each state it samples
  \texttt{num\_particles}=64 candidate actions and reweights them by a Boltzmann factor
  $\exp(Q/\lambda)$, where the temperature $\lambda$ is learned online (\texttt{temp\_lr}=$3\times10^{-4}$)
  so that the induced policy shift matches \texttt{target\_kl}=2.0 (3.0 for Hopper, 1.0 for Ant).
  The reference policy is refreshed every 1000 steps, and exploration adds Gaussian
  \texttt{additive\_noise}=0.2.

  \textit{SDAC} draws \texttt{num\_reverse\_samples}=500 Monte Carlo samples during reverse
  diffusion for gradient estimation, with temperature $\lambda$=0.05 (0.01 on DMC).

  \textit{QVPO} trains with \texttt{num\_train\_samples}=64, advantage-based reweighting,
  \texttt{entropy\_coef}=0.01, and \texttt{num\_behavior\_samples}=2 (4 for HalfCheetah, 1 for Hopper).

  \paragraph{On-policy algorithm-specific hyperparameters.}
  The four methods differ mainly in their policy parameterization. PPO uses a Gaussian actor;
  DPPO a diffusion actor; and GenPO and FPO++ continuous normalizing-flow actors. All four use
  an actor (flow/diffusion) learning rate of $10^{-4}$; the critic uses $10^{-3}$, except FPO++,
  which uses $10^{-4}$ together with a wider 3-layer $(512,512,512)$ critic. Unless noted, the
  generative actors are 3-layer $(256,256,256)$ MLPs.

  \textit{PPO} is the Gaussian baseline: a state-independent log-std initialized at
  \texttt{init\_noise\_std}=1.0, entropy bonus \texttt{entropy\_coeff}=0.005, value-loss
  coefficient $1.0$ with clipped value loss, and a fixed learning-rate schedule
  (\texttt{desired\_kl}=0.01, unused under the fixed schedule).

  \textit{DPPO} treats the $K$-step denoising chain as an MDP and runs PPO on it. The diffusion
  actor uses \texttt{steps}=10 DDPM steps (cosine schedule, Mish, \texttt{time\_dim}=32) with a
  denoising discount \texttt{gamma\_denoising}=0.99 and a per-step log-prob std floor
  \texttt{min\_logprob\_denoising\_std}=0.1. Beyond the standard clip $\epsilon=0.2$, it applies
  DPPO's per-step clipping with \texttt{clip\_epsilon\_base}=0.001 and \texttt{clip\_epsilon\_rate}=3.0.

  \textit{GenPO} optimizes a flow actor (ELU, \texttt{time\_dim}=32, \texttt{steps}=5,
  \texttt{mix\_para}=0.95) in an augmented action space, computing log-probabilities via the
  inverse-Jacobian of the flow. It adds a compression loss (\texttt{compress\_coef}=0.01) that
  pulls the two augmented action halves together, disables the entropy bonus
  (\texttt{entropy\_coeff}=0.0), and uses an adaptive learning-rate schedule targeting
  \texttt{desired\_kl}=0.01, with minibatch size $6144$.

  \textit{FPO++} performs flow policy optimization with a conditional flow-matching (CFM)
  surrogate. The flow actor uses \texttt{steps}=64 sampling steps (ELU, \texttt{time\_dim}=16,
  unclipped sampler) and a tighter clip $\epsilon=0.05$. The CFM ratio is estimated from
  \texttt{num\_mc\_samples}=16 Monte Carlo samples (32 for humanoids) over
  \texttt{num\_epochs}=16 epochs (32 for humanoids), with reduction \texttt{cfm\_loss\_reduction}=sqrt,
  inverse-CDF time sampling ($\beta=1.0$), and clamps
  \texttt{cfm\_loss\_clamp}=20, \texttt{cfm\_diff\_clamp\_max}=10, and \texttt{advantage\_clamp}=100.
  Exploration adds Gaussian \texttt{additive\_noise}=0.02; actions are clipped to
  \texttt{action\_clip}=2.0; value loss is unclipped; and an EMA of the flow weights
  (\texttt{ema\_decay}=0.95, \texttt{ema\_warmup\_steps}=500) is used for deterministic evaluation.




\begin{table}[h!]
\centering
\caption{\textbf{Reference returns used for normalization} in performance profiles (Figures~\ref{fig:mujoco_tau} and~\ref{fig:diffusion_steps}). For each environment, we select a value approximately matching the highest returns observed across the evaluated algorithms.}
\label{tab:max_returns}
\begin{tabular}{lc}
\toprule
Environment & Max Return \\
\midrule
Ant-v5 & 6000 \\
HalfCheetah-v5 & 11000 \\
Hopper-v5 & 4000 \\
Humanoid-v5 & 6000 \\
Swimmer-v5 & 150 \\
Walker2d-v5 & 6000 \\
\bottomrule
\end{tabular}
\end{table}

\end{document}